\documentclass[10pt,twocolumn,letterpaper]{article}

\usepackage{cvpr}
\usepackage{times}
\usepackage{epsfig}
\usepackage{graphicx}
\usepackage{amsmath}
\usepackage{amssymb}
\usepackage{bbm}
\usepackage{mathrsfs}
\usepackage{amsmath}
\usepackage{subfigure}
\usepackage{float}
\usepackage{booktabs}
\usepackage{multirow}
\usepackage{makecell,rotating}
\usepackage{array}
\usepackage{setspace}
\usepackage{framed}

\usepackage[pagebackref=true,breaklinks=true,letterpaper=true,colorlinks,bookmarks=false]{hyperref}

\cvprfinalcopy 

\def\cvprPaperID{6892} 

\ifcvprfinal\fi
\begin{document}

\title{Explaining Knowledge Distillation by Quantifying the Knowledge}

\author{Xu Cheng\\
Shanghai Jiao Tong University\\
{\tt\small xcheng8@sjtu.edu.cn}
\and
Zhefan Rao\\
Huazhong University of Science \& Technology\\
{\tt\small rzf19971013@gmail.com}
\and
Yilan Chen\\
Xi'an Jiaotong University\\
{\tt\small chenyilan@stu.xjtu.edu.cn}
\and
Quanshi Zhang\\
Shanghai Jiao Tong University\\
{\tt\small zqs1022@sjtu.edu.cn}
}

\maketitle

\begin{abstract}
This paper presents a method to interpret the success of knowledge distillation by quantifying and analyzing task-relevant and task-irrelevant visual concepts that are encoded in intermediate layers of a deep neural  network (DNN). More specifically, three hypotheses are proposed as follows. \textbf{1.} Knowledge distillation makes the DNN learn more visual concepts than learning from raw data. \textbf{2.} Knowledge distillation ensures that the DNN is prone to learning various visual concepts simultaneously. Whereas, in the scenario of learning from raw data, the DNN learns visual concepts sequentially. \textbf{3.}  Knowledge distillation yields more stable optimization directions than learning from raw data. Accordingly, we design three types of mathematical metrics to evaluate feature representations of the DNN. In experiments, we diagnosed various DNNs, and above hypotheses were verified.
\end{abstract}

\section{Introduction}
The success of knowledge distillation~\cite{hinton2015distilling} has been demonstrated in various studies  \cite{romero2014fitnets,yim2017gift,furlanello2018born}. It transfers knowledge from a well-learned deep neural network (DNN), namely the teacher network, to another DNN, namely the student network. However, explaining how and why knowledge distillation outperforms learning from raw data still remains a challenge.

\begin{figure}[t]
    \centering
    \includegraphics[width=\linewidth]{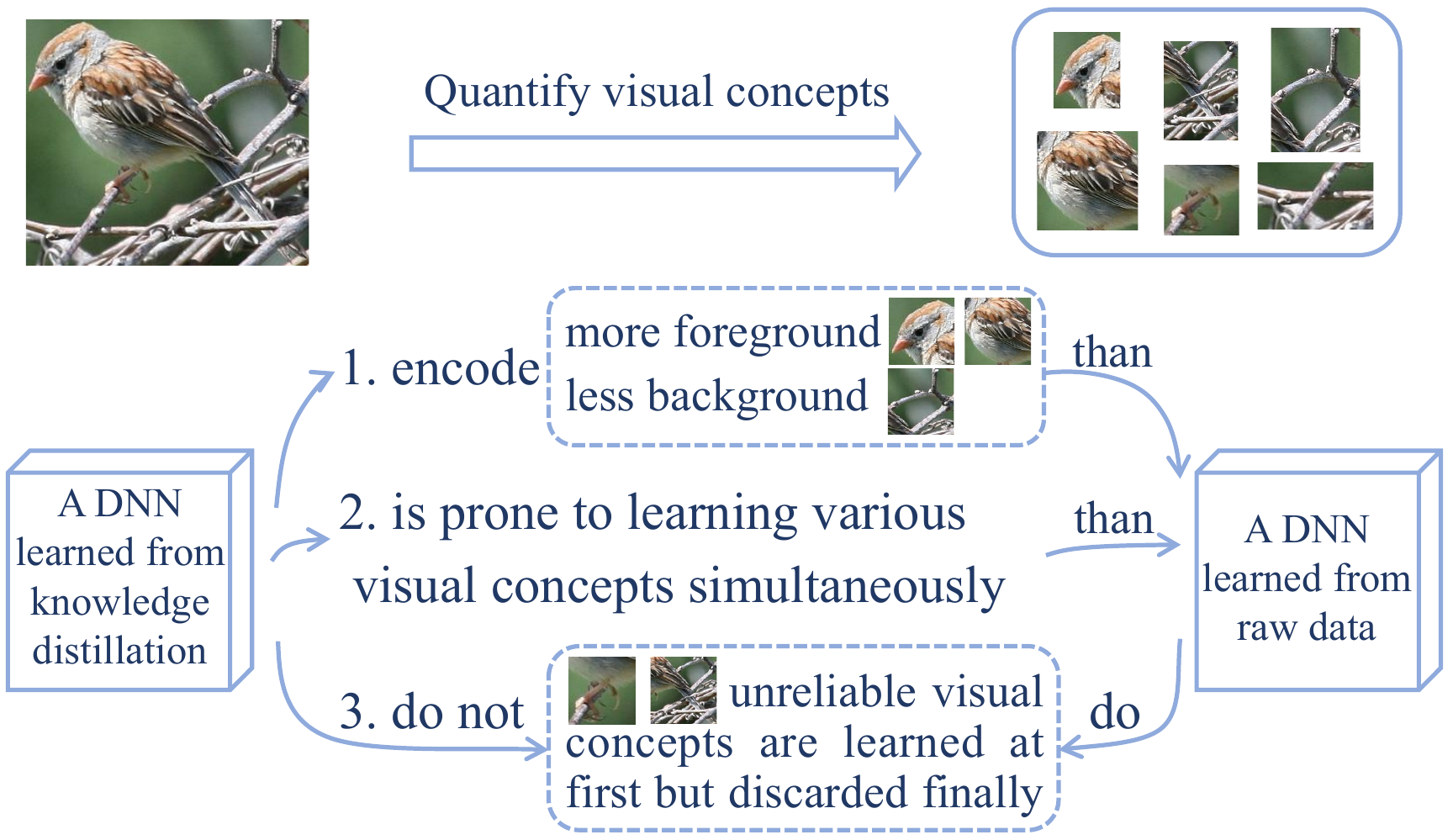}
    \caption{Explanation of knowledge distillation by quantifying visual concepts. Three hypotheses are proposed and verified as follows. 1. Knowledge distillation makes the DNN learn more visual concepts than learning from raw data. 2. Knowledge distillation ensures that the DNN is prone to learning various visual concepts simultaneously. 3. Knowledge distillation yields more stable optimization directions than learning from raw data.\vspace{-10pt}}
    \label{fig0}
\end{figure}

In this work, we aim to analyze the success of knowledge distillation from a new perspective, \emph{i.e.}~quantifying the knowledge encoded in the intermediate layer of a DNN. We quantify and compare the amount of the knowledge encoded in the DNN learned via knowledge distillation and the DNN learned from raw data, respectively. Here, the DNN learned from raw data is termed the \textbf{\emph{baseline network}}. In this research, the amount of the knowledge of a specific layer is measured as the number of visual concepts (\emph{e.g.} object parts like tails, heads and etc.), which is shown in Figure~\ref{fig0}. These visual concepts activate the feature map of this specific layer and are used for prediction.

We design three types of mathematical metrics to analyze task-relevant and task-irrelevant visual concepts. Then, these metrics are used to quantitatively verify three hypotheses as follows.

 \textbf{Hypothesis 1: Knowledge distillation makes the DNN learn more visual concepts.} In this paper, a visual concept is defined as an image region, whose information is significantly less discarded and is mainly used by the DNN. We distinguish visual concepts that are relevant to the task away from other concepts, \emph{i.e.}~task-irrelevant concepts. For implementation, let us take the classification task as an example. As is vividly shown in Figure~\ref{fig0}, visual concepts on the foreground are usually regarded task-relevant, while those on the background are considered task-irrelevant.

According to the information-bottleneck theory~\cite{wolchover2017new,shwartz2017opening}, DNNs tend to expose task-relevant visual concepts and discard task-irrelevant concepts to learn discriminative features. Compared to the baseline network (learned from raw data), a well-trained teacher network is usually considered to encode more task-relevant visual concepts and/or less task-irrelevant concepts. Because the student network mimics the logic of the teacher network, the student network is supposed to contain more task-relevant visual concepts and less task-irrelevant concepts.

\textbf{Hypothesis 2: Knowledge distillation ensures the DNN is prone to learning various visual concepts simultaneously.} In comparison, the baseline network tends to learn visual concepts sequentially, \emph{i.e.} learning different concepts in different epochs.

\textbf{Hypothesis 3: Knowledge distillation usually yields more stable optimization directions than learning from raw data.}
When learning from raw data, the DNN usually tries to model various visual concepts in early epochs and then discard non-discriminative ones in later epochs~\cite{wolchover2017new,shwartz2017opening}, which leads to unstable optimization directions. We name the phenomenon of inconsistent optimization directions through different epochs \textbf{\emph{``detours''}}\footnote[1]{``Detours'' indicate the phenomenon that a DNN tries to model various visual concepts in early epochs and discard non-discriminative ones later.} for short in this paper. In comparison, during the knowledge distillation, the teacher network directly guides the student network to target visual concepts without \emph{significant detours}\textcolor{red}{\footnotemark[1]}. Let us take the classification of birds as an example. The baseline network tends to extract features from the head, belly, tail, and tree-branch parts in early epochs, and later discards features from the tree-branch. Whereas, the student network directly learns features from the head and belly parts with less detours\textcolor{red}{\footnotemark[1]}.

\textbf{Methods:} We propose three types of mathematical metrics to quantify visual concepts hidden in intermediate layers of a DNN, and analyze how visual concepts are learned during the learning procedure. These metrics measure 1. the number of visual concepts, 2. the learning speed of different concepts, 3. the stability of optimization directions, respectively. We use these metrics to analyze the student network and the baseline network in comparative studies to prove three hypotheses. More specifically, the student network is learned via knowledge distillation, and the baseline network learned from raw data is constructed to have the same architecture as the student network.

Note that visual concepts should be quantified without manual annotations. There are mainly two reasons. 1) It is impossible for people to annotate all kinds of potential visual concepts in the world. 2) For a rigorous research, the subjective bias in human annotations should not affect the quantitative metric. To this end, \cite{guan2019towards,deepInfo} leverages the entropy to quantify visual concepts encoded in an intermediate layer.

\textbf{Contributions}: Our contributions can be summarized as follows.

1. We propose a method to quantify dark-matter concepts~\cite{xie2017learning} encoded in intermediate layers of a DNN.

2.  Based on the quantification of visual concepts, we propose three types of metrics to diagnose and interpret the superior performance of knowledge distillation from the view of knowledge representations encoded in a DNN.

3. Three hypotheses about knowledge distillation are proposed and verified, which shed light on the explanation of the knowledge distillation.

\section{Related Work}
Although deep neural networks have exhibited superior performance in various tasks, they are still regarded as black boxes. Previous studies of interpreting DNNs can be roughly summarized into semantic explanations and mathematical explanations of the representation capacity.

\textbf{Semantic explanations for DNNs:} An intuitive way to interpret DNNs is to visualize visual concepts encoded in intermediate layers of DNNs. Feature visualization methods usually show concepts that may significantly activate a specific neuron of a certain layer. Gradient-based methods~\cite{zeiler2014visualizing,simonyan2017deep,yosinski2015understanding,mahendran2015understanding} used gradients of outputs \emph{w.r.t.} the input image to measure the importance of intermediate-layer activation units or input units. Inversion-based~\cite{dosovitskiy2016inverting} methods inverted feature maps of convolutional  layers into images. From visualization results, people roughly understand visual concepts encoded in intermediate layers of DNNs. For example, filters in low layers usually encode simple visual concepts such as edges and textures, and filters in high layers usually encode concepts \emph{e.g.} objects and patterns.

Other methods usually estimated the pixel-wise attribution/importance/saliency on an input image, which measured the influence of each input pixel to the final output~\cite{ribeiro2016should,lundberg2017unified,kindermans2017learning,fong2017interpretable}.
Some methods explored the saliency of the input image using intermediate-layer features, such as CAM~\cite{zhou2016learning}, Grad-CAM~\cite{selvaraju2017grad}, and Grad-CAM++~\cite{chattopadhay2018grad}. Zhou~\etal~\cite{zhou2014object} computed the actual image-resolution receptive field of neural activations in a feature map.

Bau~\etal~\cite{bau2017network} disentangled feature representations into semantic concepts using human annotations. Fong and Vedaldi~\cite{fong2018net2vec} demonstrated that a DNN used multiple filters to represent a specific semantic concept. Zhang~\etal used an explanatory graph~\cite{explanatoryGraph} and a decision tree~\cite{explanatoryTree} to represent hierarchical compositional part representations in CNNs. TCAV~\cite{kim2017interpretability} measured the importance of user-defined concepts to classification.

Another direction of explainable AI is to learn a DNN with interpretable feature representations in an unsupervised or weakly-supervised manner. In the capsule network~\cite{sabour2017dynamic}, activities of each capsule encoded various properties. The interpretable CNN~\cite{interpretableCNN} learned object part features without part annotations. InfoGAN~\cite{chen2016infogan} and $\beta$-VAE~\cite{higgins2017beta} learned interpretable factorised latent representations for generative networks.

In contrast, in this research, the quantification of intermediate-layer visual concepts requires us to design metrics with coherency and generality. \emph{I.e.} unlike previous studies compute importance/saliency/attention~\cite{zeiler2014visualizing,simonyan2017deep,yosinski2015understanding,mahendran2015understanding} based on heuristic assumptions or using massive human-annotated concepts~\cite{bau2017network} to explain network features, we quantify visual concepts using the conditional entropy of the input. The entropy is a generic tool with strong connections to various theories,~\emph{e.g.} the information-bottleneck theory~\cite{wolchover2017new,shwartz2017opening}. Moreover, the coherency allows the same metric to ensure fair comparisons between layers of a DNN, and between DNNs learned in different epochs.

\textbf{Mathematical explanations for the representation capacity of DNNs:} Evaluating the representation capacity of DNNs mathematically provides a new perspective for explanations. The information-bottleneck theory~\cite{wolchover2017new,shwartz2017opening} used the mutual information to evaluate the representation capacity of DNNs~\cite{goldfeld2019estimating,xu2017information}. The stiffness~\cite{fort2019stiffness} was proposed to diagnose the generalization of a DNN. The CLEVER score~\cite{weng2018evaluating} was used to estimate the robustness of neural networks. The Fourier analysis~\cite{xu2018understanding} was applied to explain the generalization of DNNs learned by stochastic gradient descent. Novak \etal ~\cite{novak2018sensitivity} investigated the correlation between the sensitivity of trained neural networks and generalization. Canonical correlation analysis (CCA) ~\cite{kornblith2019similarity} was used to measure the similarity between representations of neural networks. Chen~\etal~\cite{chen2018learning} proposed instance-wise feature selection via mutual information for model interpretation. Zhang~\etal~\cite{consistency} explored knowledge consistency between DNNs.

Different from previous methods, our research aims to bridge the gap between mathematical explanations and semantic explanations. We use the entropy of the input to measure the number of visual concepts in a DNN. Furthermore, we quantify visual concepts on the background and the foreground \emph{w.r.t.} the input image, explore whether a DNN learn various concepts simultaneously or sequentially, and analyze the stability of optimization directions.

\textbf{Knowledge distillation:} knowledge distillation is a popular and successful technique in knowledge transferring. Hinton
~\etal~\cite{hinton2015distilling} considered  ``soft targets'' led to the superior performance of knowledge distillation.  Furlanello~\etal ~\cite{furlanello2018born} explained the dark knowledge transferred from the teacher to the student as importance weighting.

From a theoretical perspective, Lopez-Paz~\etal~\cite{lopez2015unifying} interpreted knowledge distillation as a form of learning with privileged information. Phuong~\etal~\cite{phuong2019towards} explained the success of knowledge distillation from the view of data distribution, optimization bias, and the size of the training set.

However, to the best of our knowledge, the mathematical explanations for knowledge distillation are rare. In this paper, we interpret knowledge distillation from a new perspective, \emph{i.e.} quantifying, analyzing, and comparing visual concepts encoded in intermediated layers between DNNs learned by knowledge distillation and DNNs learned purely from raw data mathematically.

\section{Algorithm}
In this section, we are given a pre-trained DNN (\emph{i.e.} the teacher network) and then distill it into another DNN (\emph{i.e.} the student network). In this way, we aim to compare and explain the difference between the student network and the DNN learned from raw data (\emph{i.e.} the baseline network). To simplify the story, we limit our attention to the task of object classification. Let  $x\in{R^{n}}$ denote the input image, and $f_{T}(x),f_{S}(x)\in{R^{L}}$ denote intermediate-layer features of the teacher network and its corresponding student network, respectively. Knowledge distillation is conducted to force $f_{S}(x)$ to approximate $f_{T}(x)$. Classification results of the teacher and the student are given as $y_{T}=g_{T}(f_{T}(x))$ and  $y_{S}=g_{S}(f_{S}(x))\in{R^{c}}$, respectively.

We compare visual concepts encoded in the baseline network and those in the student network to explain knowledge distillation. For a fair comparison, the baseline network has the same structure as the student network, and implementation details are shown in Section~\ref{Implementation Details}.

\subsection{Preliminaries: Quantification of Information Discarding}
According to the information-bottleneck theory~\cite{wolchover2017new,shwartz2017opening}, the information of the input image is gradually discarded through layers. \cite{guan2019towards,deepInfo} proposed a method to quantify the input information that was encoded in a specific intermediate layer of a DNN, \emph{i.e.}~measuring how much input information was neglected when the DNN extracted the feature of this layer. The information discarding is formulated as the conditional entropy $H(X')$ of the input, given the intermediate-layer feature $f^{*}=f(x)$,~as follows.
\begin{equation}\label{1}
\small{
H(X')   ~~~\emph{s.t.}  ~~\forall{x'\in{X'},} ~~\parallel{f(x')-f^{*}}\parallel ^{2} \le \tau}
\end{equation}
$X'$ denotes a set of images which correspond to the concept of a specific object instance. The concept of the object is assumed to be represented by a small range of features $\small{ \parallel{f(x')-f^{*}}\parallel ^{2}\le \tau}$, where $\tau$ is a small positive scalar. It was assumed that $x'$ follows an ${i.i.d.}$ Gaussian distribution, $x' \sim \mathcal{N}(x,\Sigma = diag(\sigma_{1}^{2},\dots, \sigma_{n}^{2}))$, where $\sigma_{i}$ controls the magnitude of the perturbation at each $i$-th pixel. $n$ denotes the number of pixels of the input image. In this way, the assumption of the Gaussian distribution ensures that the entropy $H(X') $ of the entire image can be decomposed into pixel-level entropies $\{H_{i}\}$ as follows.
\begin{equation}\label{2}
\small{
H(X') = \sum_{i=1}^{n} H_{i}}
\end{equation}
where $H_{i} = \log\sigma_{i} +  \frac{1}{2} \log(2\pi{e})$ measures the discarding of pixel-wise information. Please see \cite{guan2019towards,deepInfo} for details.


\subsection{Quantification of visual concepts}
\begin{framed}
\vspace{-5pt}
\label{Quantification of visual concepts}
\textbf{Hypothesis 1}: Knowledge distillation makes the DNN learn more reliable visual concepts than learning from raw data.
\vspace{-5pt}
\end{framed}\vspace{-6pt}

In this section, we aim to compare the number of visual concepts that are encoded in the baseline network and those in the student network, so as to verify the above hypothesis.

\textbf{Using annotated concepts or not: }For comparison, we try to define and quantify visual concepts encoded in the intermediate layer of a DNN (either the student network or the baseline network). Note that, in this study, we do \textbf{not} study visual concepts defined by human annotations. For example, Bau~\etal~\cite{bau2017network} defined visual concepts of objects, parts, textures, scenes, materials, and colors by using manual annotations. However, this research requires us to use and quantify visual concepts without explicit names, which cannot be accurately labeled. These visual concepts are usually referred to as \emph{\textbf{``Dark Matters'' }}~\cite{xie2017learning}.

There are mainly two reasons to use dark-matter visual concepts instead of traditional semantic visual concepts. 1. There exist no standard definitions for semantic visual concepts, and the taxonomy of semantic visual concepts may have significant bias. 2. The cost of annotating all visual concepts are usually unaffordable.

\begin{figure}[t]
    \centering
    \includegraphics[width=0.9\linewidth]{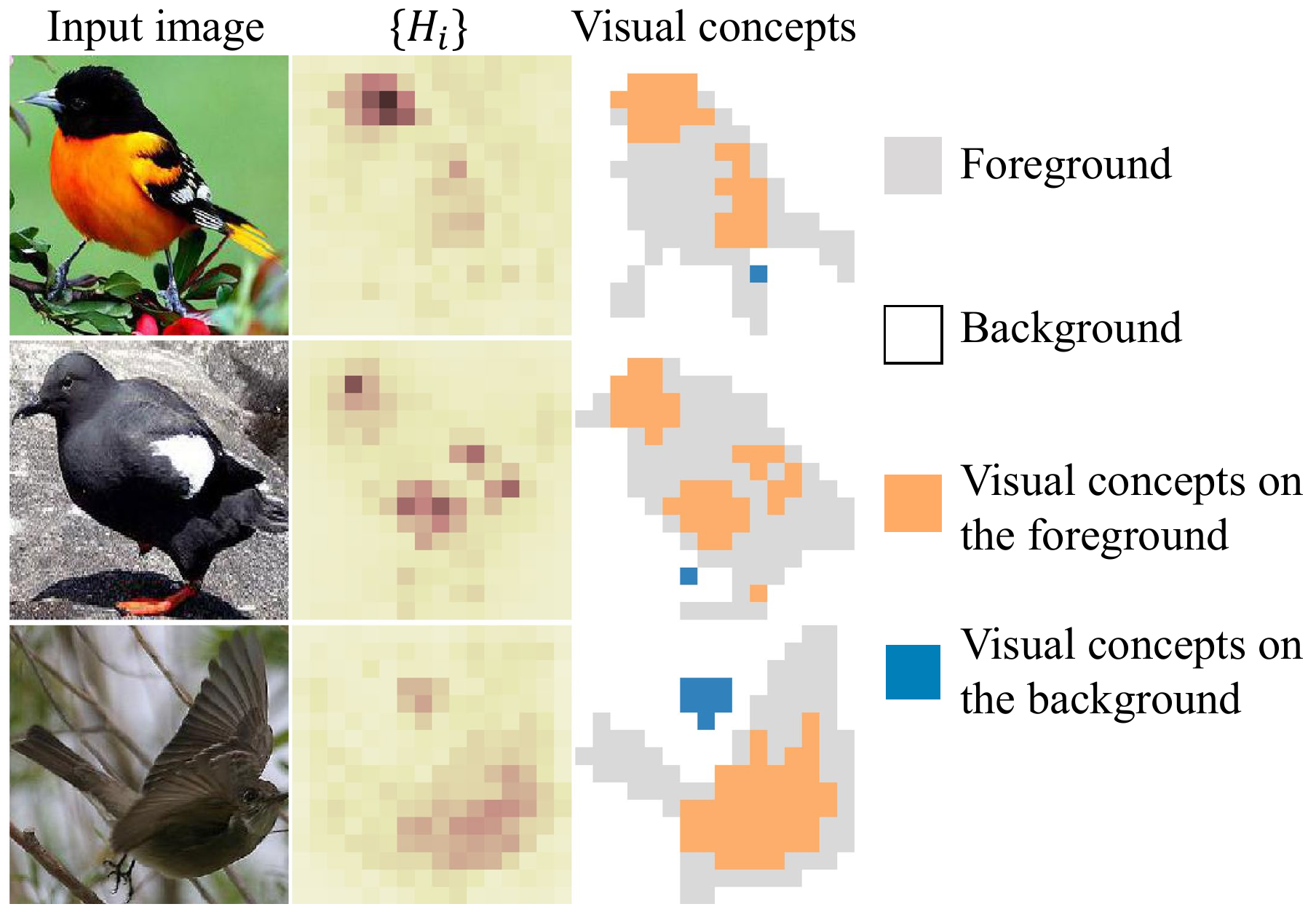}
    \caption{Visualization of visual concepts. The second column shows $\{H_{i}\}$ of different images. Image regions with low pixel-wise entropies ${\{H_{i}\}}$ are considered as visual concepts, which are shown in the third column.\vspace{-10pt}}
    \label{fig1}
\end{figure}

\textbf{Metric:} In this paper, we quantify dark-matter visual concepts from the perspective of the information theory. Given a pre-trained DNN, a set of training images $\textbf{I}$ and an input image $x \in \textbf{I} $, let us focus on the pixel-wise information discarding ${H_{i}}$ \emph{w.r.t.} the intermediate-layer feature $f^{*}=f(x)$. High pixel-wise entropies ${\{H_{i}\}}$, shown in Equation \eqref{2}, indicate that the DNN neglects more information of these pixels. Whereas, the DNN mainly utilizes pixels with low entropies ${\{H_{i}\}}$ to compute the feature $f^{*}$. In this way, image regions with low pixel-wise entropies ${\{H_{i}\}}$ can be considered to represent relatively valid visual concepts. For example, the head and wings of the bird in Figure \ref{fig1} are mainly used by the DNN for fine-grained classification. Therefore, metrics are defined as follows.
\begin{equation}\label{3}
\begin{small}
\begin{aligned}
\begin{split}
&N_{\textrm{concept}}^{\textrm{bg}}(x) =\sum_{i \in \Lambda_{\textrm{bg}}~\textrm{\emph{w.r.t.} }x} \mathbbm{1}(\overline{H} - H_{i} > b),\\
&N_{\textrm{concept}}^{\textrm{fg}}(x) =\sum_{i \in \Lambda_{\textrm{fg}}~\textrm{\emph{w.r.t.} }x} \mathbbm{1}(\overline{H} - H_{i} > b),\\
& \lambda = \mathbb{E}_{x \in \textbf{I}}[N_{\textrm{concept}}^{\textrm{fg}}(x) / (N_{\textrm{concept}}^{\textrm{fg}}(x) + N_{\textrm{concept}}^{\textrm{bg}}(x))]
\end{split}
\end{aligned}
\end{small}
\end{equation}
where $N_{\textrm{concept}}^{\textrm{bg}}(x)$ and  $N_{\textrm{concept}}^{\textrm{fg}}(x)$ denote the number of visual concepts encoded on the background and the foreground, respectively. $\Lambda_{\textrm{bg}}$ and $\Lambda_{\textrm{fg}}$ are sets of pixels on the background and the foreground \emph{w.r.t.} the input image $x$, respectively.
$\mathbbm{1}(\cdot)$ is the indicator function. If the condition inside is valid, $\mathbbm{1}(\cdot)$ returns $1$, and otherwise $0$.~$\overline{H}=\mathbb{E}_{i \in \Lambda_{\textrm{bg}}}[H_{i}]$ denotes the average entropy value of the background, which measures the significance of information discarding~\emph{w.r.t.} background pixels. Those pixels on the background are considered to represent task-irrelevant visual concepts. Therefore, we can use $\overline{H}$ as a baseline entropy. Image regions with significantly lower entropy values than $\overline{H}$ can be considered as valid visual concepts, where $b$ is a positive scalar. The metric $\lambda$ is used to measure the discriminative power of features. As shown in Figure \ref{fig1}, in order to improve the stability and efficiency of the computation, ${\{H_{i}\}}$ is computed in $16 \times 16$ grids, \emph{i.e.} all pixels in each local grid share the same $\sigma_{i}$. The dark color in Figure \ref{fig1} indicates the low entropy value ${H_{i}}$.

In statistics, visual concepts on the foreground are usually task-relevant, while those on the background are mainly task-irrelevant. In this way, a well-learned DNN is supposed to encode a large number of visual concepts on the foreground and very few ones on the background. Thus, a larger $\lambda$ value denotes the DNN is more discriminative.

\textbf{Generality and coherency}: The design of a metric should consider both the generality and the coherency. The generality is referred to as that the metric is supposed to have strong connections to existing mathematical theories. The coherency ensures comprehensive and fair comparisons in different cases. In this paper, we aim to quantify and compare the number of visual concepts between different network architectures and between different layers. As discussed in \cite{guan2019towards,deepInfo}, existing methods of explaining DNNs usually depend on specific network architecture or specific tasks, such as gradient-based methods~\cite{zeiler2014visualizing,simonyan2017deep,yosinski2015understanding,mahendran2015understanding}, perturbation-based methods~\cite{fong2017interpretable,kindermans2017learning} and inversion-based methods~\cite{dosovitskiy2016inverting}. Unlike previous methods, the conditional entropy of the input ensures fair comparisons between different network architectures and between different layers, which is reported in Table~\ref{Generality and coherency}.

\subsection{Learning simultaneously or sequentially}
\begin{framed}
\vspace{-5pt}
\textbf{Hypothesis 2}: Knowledge distillation ensures that the DNN is prone to learning various concepts simultaneously. Whereas, the DNN learned from raw data learns concepts sequentially through different epochs.
\vspace{-5pt}
\end{framed}

\begin{table}[t]
\begin{center}
\resizebox{0.9\linewidth}{!}{\
\begin{small}
\begin{tabular}{c| c|c| c}
\toprule
\multirow{2}*{} &\multirow{2}*{Generality}& \multicolumn{2}{c}{Coherency} \\
\cline{3-4}
& & Layers & Nets  \\
\midrule
Gradient-based~\cite{zeiler2014visualizing,simonyan2017deep,yosinski2015understanding,mahendran2015understanding} & No& No& No\\
\midrule
Perturbation-based~\cite{fong2017interpretable,kindermans2017learning} & No& No& No\\
\midrule
Inversion-based~\cite{dosovitskiy2016inverting}  & No& No& No\\
\midrule
Entropy-based & Yes& Yes& Yes\\
\bottomrule
\end{tabular}
\end{small}}
\end{center}
\caption{Comparisons of different methods in terms of generality and coherency. The entropy-based method provides coherent results across layers and networks.}
\label{Generality and coherency}
\end{table}

In this section, We propose two metrics to prove Hypothesis 2. Given a set of training images $\textbf{I}$, $g_{1}, g_{2}, \dots, g_{M}$ denote DNNs learned in different epochs. This DNN can be either the student network or the baseline network. $g_{M}$ obtained after the last epoch $M$ is regarded as the final DNN. For each specific image $I\in{\textbf{I}}$, we quantify visual concepts on the foreground encoded in DNNs learned after different epochs $N_{1}^{\textrm{fg}}(I), N_{2}^{\textrm{fg}}(I),\dots,  N_{M}^{\textrm{fg}}(I)$.

In this way, whether or not a DNN learns visual concepts simultaneously can be analyzed in following two terms: 1. whether $N_{j}^{\textrm{fg}}(I)$ increases fast along with the epoch number; 2. whether $N_{j}^{\textrm{fg}}(I)$ of different images increases simultaneously. The first term indicates whether a DNN learns various visual concepts of a specific image quickly, while the second term evaluates whether a DNN learns visual concepts of different images simultaneously.

\begin{figure}[t]
    \centering
    \includegraphics[width=0.7\linewidth]{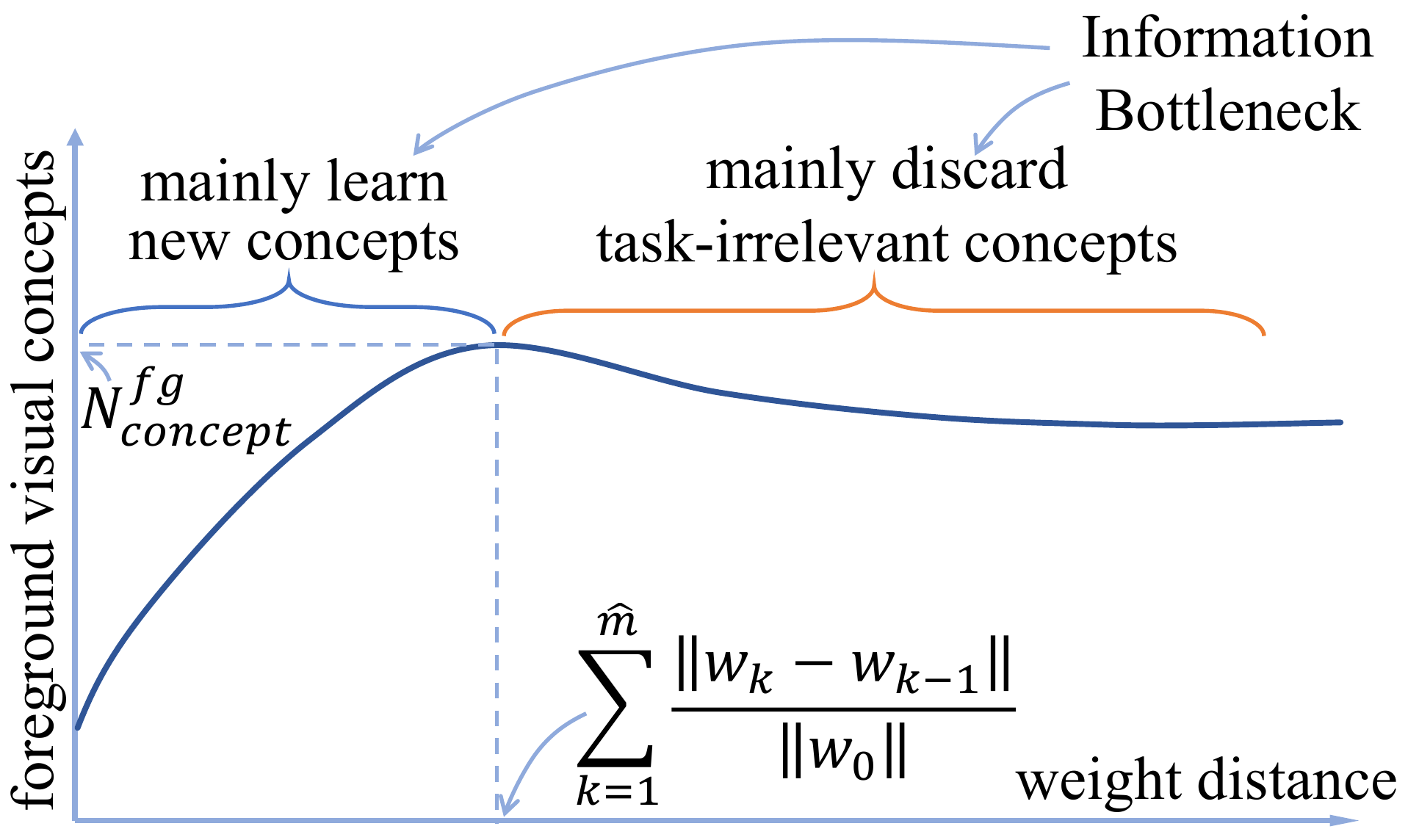}
    \caption{Procedure of learning foreground visual concepts \emph{w.r.t.} weight distances. According to the information-bottleneck theory, a DNN tends to learn various visual concepts mainly in early stages and then mainly discard task-irrelevant concepts later. Strictly speaking, a DNN learns new concepts and discards old concepts during the entire stages. We can consider that the learning stage of $\hat{m}$ encodes richest concepts.\vspace{-10pt}}
    \label{fig2}
\end{figure}

For a rigorous evaluation, as shown in Figure \ref{fig2}, we calculate the epoch number $\hat{m} = \arg\max_{k} N_{k}^{\textrm{fg}}(I)$, where a DNN obtains richest visual concepts on the foreground. Let $w_{0}$ and $w_{k}$ denote initial parameters and parameters learned after the $k$-th epoch. We utilize $\sum_{k=1}^{\hat{m}} \frac{\|{w_{k}-w_{k-1}}\|}{\|{w_{0}}\|}$, named \textbf{\emph{``weight distance'' }}, to measure the learning effect at $\hat{m}$-th epoch~\cite{garipov2018loss,flennerhag2018transferring}. Compared to using the epoch number, the weight distance better quantifies the total path of updating the parameter $w_{k}$ at each epoch $k$. Thus, we use the average value $D_{\textrm{mean}}$ and standard deviation value $D_{\textrm{std}}$ of weight distances to quantify whether a DNN learns visual concepts simultaneously. $D_{\textrm{mean}}$ and $D_{\textrm{std}}$ are given as follows.
\begin{equation}\label{4}
\small{
\begin{split}
D_{\textrm{mean}} =\underset{I\in \textbf{I}}{\mathbb{E}}\bigg[ \sum_{k=1}^{\hat{m}} \frac{\|{w_{k}-w_{k-1}}\|}{\|{w_{0}}\|}\bigg],\\
D_{\textrm{std}} = \underset{I\in \textbf{I}}{Var}\bigg[\sum_{k=1}^{\hat{m}} \frac{\|{w_{k}-w_{k-1}}\|}{\|{w_{0}}\|}\bigg]
\end{split}}
\end{equation}
$D_{\textrm{mean}}$ represents the average weight distance, where the DNN obtains the richest task-relevant visual concepts.  The value of $D_{\textrm{mean}}$ indicates whether a DNN learns visual concepts quickly. $D_{\textrm{std}}$ describes the variation of the weight distance \emph{w.r.t} different images, and its value denotes whether a DNN learns various visual concepts simultaneously. Hence, small values of $D_{\textrm{mean}}$ and $D_{\textrm{std}}$ indicate that the DNN learns various concepts quickly and simultaneously.

\subsection {Learning with Less Detours}
\begin{framed}
\vspace{-5pt}
\textbf{Hypothesis 3}: Knowledge distillation yields more stable optimization directions than learning from raw data.
\vspace{-5pt}
\end{framed}\vspace{-6pt}

During the knowledge distillation, the teacher network directly guides the student network to learn target visual concepts without significant detours\textcolor{red}{\footnotemark[1]}. In comparison, according to the information-bottleneck theory~\cite{wolchover2017new,shwartz2017opening}, when learning from raw data, the DNN usually tries to model various visual concepts and then discard non-discriminative ones, which leads to unstable optimization directions.

\begin{figure}[t]
    \centering
    \includegraphics[width=0.86\linewidth]{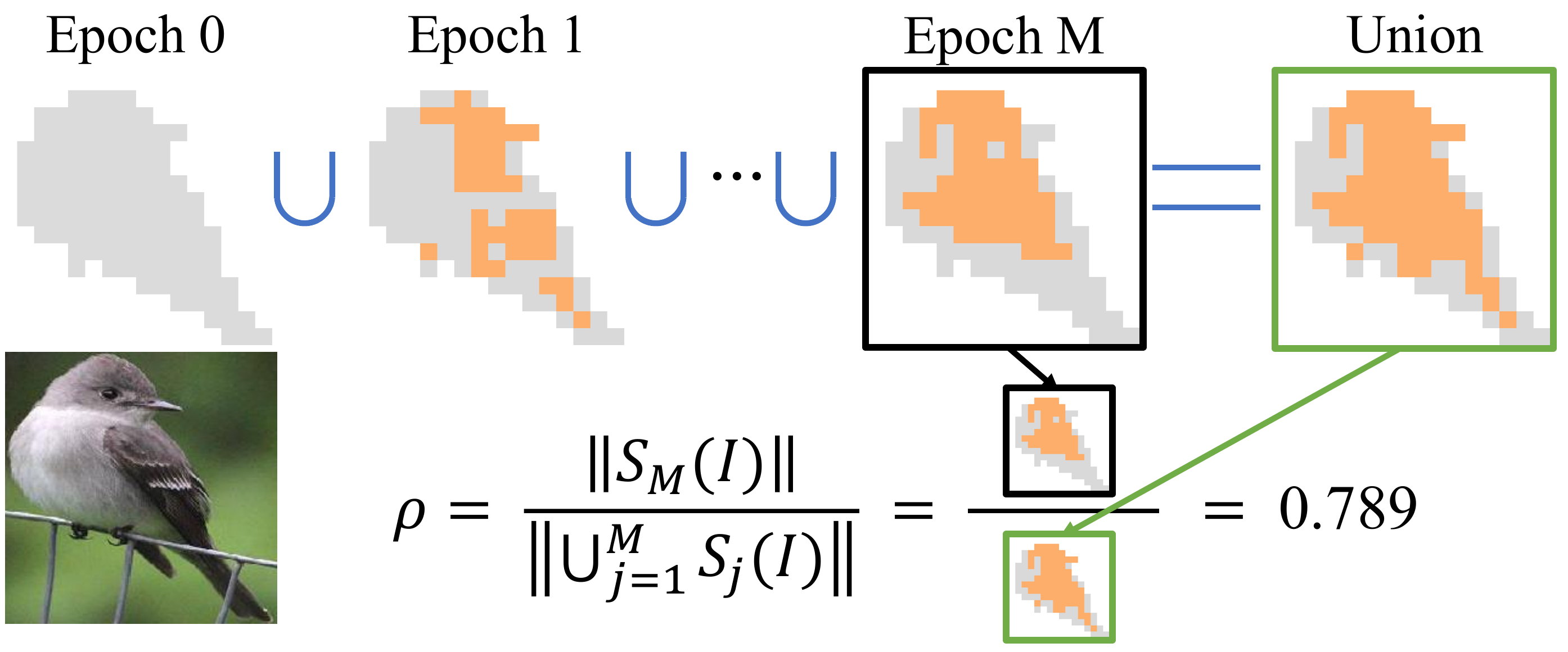}
    \caption{Detours\protect\footnotemark[1] of learning visual concepts. We visualize sets of foreground visual concepts learned after different epochs. The green box indicates the union of visual concepts learned during all epochs. The $(1-\rho)$ value denotes the ratio of visual concepts that are discarded during the learning procedure \emph{w.r.t.} the union set of concepts. Thus, a larger $\rho$ value indicates that the DNN learns with less detours\protect\footnotemark[1].\vspace{-10pt}}
    \label{fig3}
\end{figure}

In order to quantify the stability of optimization directions of a DNN, a new metric is proposed.  Let $S_{1}(I), S_{2}(I), \dots,  S_{M}(I)$ denote the set of visual concepts on the foreground of image $I$ encoded by $g_{1}, g_{2}, \dots, g_{M}$, respectively. Here, each visual concept $a \in{S_{j}(I)}$ is referred to as a specific pixel $i$ on the foreground of image $I$, which satisfies $\overline{H} - H_{i} > b$. The stability of optimization directions can be measured as follows.
\begin{equation}\label{5}
\begin{small}
 \rho = \frac{\|S_{M}(I)\|}{\|\bigcup_{j=1}^{M} S_{j}(I)\|}
 \end{small}
\end{equation}
The numerator reflects the number of visual concepts, which have been chosen ultimately for object classification, and are shown as the black box in Figure \ref{fig3}. The denominator represents visual concepts temporarily learned during the learning procedure, which is shown as the green box in Figure \ref{fig3}. $(\bigcup_{j=1}^{M} S_{j}(I)\setminus{S_{M}(I))}$ denotes the set of visual concepts, which have been tried, but finally are discarded by the DNN. A high value of $\rho$ indicates that the DNN is optimized with less detours\textcolor{red}{\footnotemark[1]} and more stably; vice versa.

\section{Experiment}
\subsection{Implementation Details}
\label{Implementation Details}
\textbf{Datasets \& DNNs:} We designed comparative experiments to verify three proposed hypotheses. For comprehensive comparisons, we conducted experiments based on AlexNet~\cite{krizhevsky2012imagenet}, VGG-11, VGG-16, VGG-19~\cite{simonyan2015very}, ResNet-50, ResNet-101, and ResNet-152~\cite{he2016deep}. Given each DNN as the teacher network, we distilled knowledge from the teacher network to the student network, which had the same architecture as the teacher network for fair comparisons. Meanwhile, the baseline network was also required to have the same architecture as the teacher network.

We trained these DNNs based on the ILSVRC-2013 DET dataset~\cite{sermanet2013overfeat}, the CUB200-2011 dataset~\cite{wah2011caltech}, and the Pascal VOC 2012 dataset~\cite{Everingham10}. All teacher networks in Sections \ref{Verification of Hypothesis 1},~\ref{Verification of Hypothesis 2},~\ref{Verification of Hypothesis 3} were pre-trained on the ImageNet dataset~\cite{russakovsky2015imagenet}, and then fine-tuned using all three datasets, respectively. For fine-tuning on the ILSVRC-2013 DET dataset, we conducted the classification of terrestrial mammal categories for comparative experiments, considering the high computational burden. For the ILSVRC-2013 DET dataset and the Pascal VOC 2012 dataset, data augmentation~\cite{jacobsen2018revnet} was applied to prevent overfitting. For the CUB200-2011 dataset, we used object images cropped by object bounding boxes for both training and testing. In particular, for the Pascal VOC 2012 dataset, images were cropped by using $1.2 \,{width} \times 1.2 \,{height}$ of the original object bounding box for stable results. For the ILSVRC-2013 DET dataset, we cropped each image by using $1.5 \,{width} \times 1.5 \,{height}$ of the original object bounding box. Because there existed no ground-truth annotations of object segmentation in the ILSVRC-2013 DET dataset, we used the object bounding box as the foreground region. Pixels within the object bounding box were regarded as the foreground~$\Lambda_{\textrm{fg}}$, and pixels outside the object bounding box were referred to as the background~$\Lambda_{\textrm{bg}}$.

\textbf{Distillation:} In the procedure of knowledge distillation, we selected a fully-connected (FC) layer $l$ as the target layer. $\|f_{T}(x)-f_{S}(x)\|^{2}$ was used as the distillation loss to mimic the feature of the corresponding layer of the teacher network, where $f_{T}(x)$ and $f_{S}(x)$ denoted the $l^{th}$-layer features of the teacher network and its corresponding student network, respectively. 

Parameters of the student network under the target FC layer $l$ were learned exclusively using the distillation loss. Hence, the learning process was not affected by the information of additional human annotations except the knowledge encoded in the teacher network, which ensured fair comparisons. Then we froze parameters under the target layer $l$ and learned parameters above the target layer $l$ merely using the classification loss.

\textbf{Selection of layers:} For each pair of the student network and the baseline network, we aimed to quantify visual concepts encoded in FC layers and thus conducted comparative experiments. We found that these selected DNNs usually had three FC layers. For sake of brevity, we named three FC layers $\textrm{FC}_{1}, \textrm{FC}_{2}, \textrm{FC}_{3}$ for short, respectively. Note that, for the ILSVRC-2013 DET dataset and the Pascal VOC 2012 dataset, dimensions of intermediate-layer features encoded in the $ \textrm{FC}_{3}$ layer were much smaller than feature dimensions of  the $\textrm{FC}_{1}$ and $ \textrm{FC}_{2}$ layers. Hence, the target layer was chosen from the $\textrm{FC}_{1}$ and $ \textrm{FC}_{2}$ layer, when DNNs were learned on the ILSVRC-2013 DET dataset and the Pascal VOC 2012 dataset. For the CUB200-2011 dataset, all three FC layers were selected as target layers. Note that, ResNets usually only had one FC layer. In this way, we replaced the only FC layer into a block with two convolutional and three FC layers, each followed by a ReLU layer. Thus, we could measure visual concepts in the student network and the baseline network \emph{w.r.t.} each FC layer. For the hyper-parameter $b$ (shown in Equation \eqref{3}), it was set to $0.25$ for AlexNet, and was set to $0.2$ for other DNNs. It was because AlexNet had much less layers than other DNNs.

\subsection{Quantification of Visual Concepts in the Teacher Network, the Student Network and the Baseline Network}
\label{4.2}
According to our hypotheses, the teacher network was learned from a large number of training data. Hence, the teacher network had learned better representations, \emph{i.e.} encoding more visual concepts on the foreground and less concepts on the background than the baseline network. Thus, the student network learned from the teacher was supposed to contain more visual concepts on the foreground than the baseline network. In this section, we aimed to compare the number of visual concepts encoded in the teacher network, the student network, and the baseline network.

We learned a teacher network from scratch, on the ILSVRC-2013 DET dataset and the CUB200-2011 dataset. In order to boost the performance of the teacher network,  data augmentation~\cite{jacobsen2018revnet} was used. The student network was distilled in the same way as Section \ref{Implementation Details}, which had the same architecture as the teacher network and the baseline network. Without loss of generality, VGG-16 was chosen, and results were reported in Table~\ref{teacher supplementary}. We found that the number of concepts on the foreground $N_{\textrm{concept}}^{\textrm{fg}}$ and the ratio $\lambda$ of the teacher network were larger than those of the student network. Meanwhile, the student network obtained larger values of $N_{\textrm{concept}}^{\textrm{fg}}$ and $\lambda$ than the baseline network. In this way, the assumed relationship between the teacher network, the student network, and the baseline network was roughly verified. We also noticed that there was an exception that the $N_{\textrm{concept}}^{\textrm{fg}}$ value of the teacher network was smaller than that of the student network. It was because the teacher network had a larger average background entropy value $\overline{H}$ (in Equation \eqref{3}) than the student network.


\begin{table}[t]
\begin{center}
\resizebox{0.9\linewidth}{!}{\
\begin{small}
\begin{tabular}{p{0.2\linewidth}<{\centering}| p{0.2\linewidth}<{\centering} p{0.1\linewidth}<{\centering}| p{0.2\linewidth}<{\centering} p{0.1\linewidth}<{\centering}}
\toprule
Dataset &Layer& &$N_{\textrm{concept}}^{\textrm{fg}}\uparrow$&  $\lambda\uparrow$\\
\midrule
  \multirow{9}*{CUB}&\multirow{3}*{\footnotesize{VGG-16 $\textrm{FC}_{1}$}}&{T} & \textbf{34.00}  & \textbf{0.78}\\
& &{S} & {29.57}  & {0.75} \\
& &{B} & 22.50 & 0.68 \\
\cline{2-5}
 &\multirow{3}*{\footnotesize{VGG-16 $\textrm{FC}_{2}$}}&T & \textbf{34.62}  & \textbf{0.80} \\
& &S & {32.92}  & {0.75} \\
& &B & 23.31 & 0.67 \\
\cline{2-5}
&\multirow{3}*{\footnotesize{VGG-16 $\textrm{FC}_{3}$}}&T & \textbf{33.97}  & \textbf{0.81}\\
& &S & {29.78}  & 0.63 \\
& &B & {23.26} & {0.71} \\
\midrule

 \multirow{6}*{ILSVRC}&\multirow{3}*{\footnotesize{VGG-16 $\textrm{FC}_{1}$}}&T & \textbf{36.80} & \textbf{0.87} \\
& &S & {35.98} & {0.84} \\
& &B & {36.47}& 0.81 \\
\cline{2-5}
 &\multirow{3}*{\footnotesize{VGG-16 $\textrm{FC}_{2}$}}&T & 38.76 & \textbf{0.89} \\
& &S & \textbf{42.74} & {0.82}\\
& &B & {36.35} & 0.82 \\

\bottomrule
\end{tabular}
\end{small}}
\end{center}
\caption{Comparisons of visual concepts encoded in the teacher network (T), the student network (S) and the baseline network (B). The teacher network encoded more visual concepts on the foreground $N_{\textrm{concept}}^{\textrm{fg}}$ and obtained a larger ratio $\lambda$ than the student network. Meanwhile, the student network had larger values of $N_{\textrm{concept}}^{\textrm{fg}}$ and $\lambda$ than the baseline network.\vspace{-10pt}}
\label{teacher supplementary}
\end{table}

\subsection{Verification of Hypothesis 1}
\label{Verification of Hypothesis 1}
Hypothesis 1 assumed that knowledge distillation ensured the student network to learn more task-relevant visual concepts and less task-irrelevant visual concepts. Thus, we utilized $N_{\textrm{concept}}^{\textrm{fg}}$, $N_{\textrm{concept}}^{\textrm{bg}}$ and $\lambda$ metrics in Equation \eqref{3} to verify this hypothesis.

\begin{figure*}[t]
    \centering
    \includegraphics[width=0.95\linewidth]{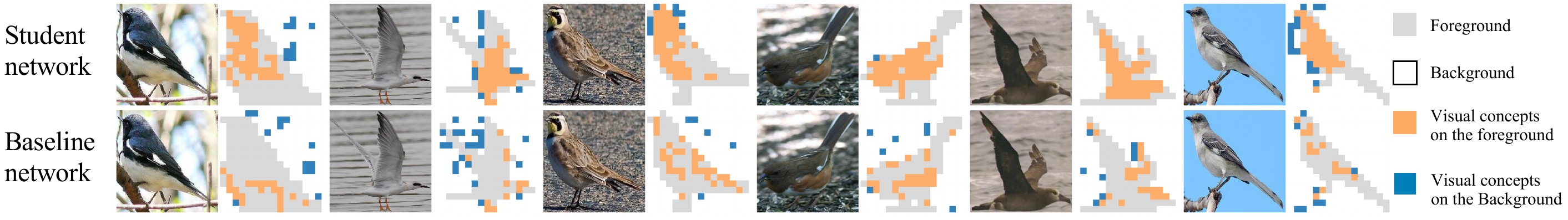}
    \caption{Visualization of visual concepts encoded in the $ \textrm{FC}_{1}$ layer of VGG-11. Generally, the student network  had a larger $N_{\textrm{concept}}^{\textrm{fg}}$ value and a smaller $N_{\textrm{concept}}^{\textrm{bg}}$ value than the baseline network.\vspace{-10pt}}
    \label{fig4}
\end{figure*}

Values of $N_{\textrm{concept}}^{\textrm{fg}}$, $N_{\textrm{concept}}^{\textrm{bg}}$ and $\lambda$, which evaluated at the $\textrm{FC}_{1}$ and $ \textrm{FC}_{2}$ layers of each DNN learned using the CUB200-2011 dataset, the ILSVRC-2013 dataset and the Pascal VOC 2012 dataset, were shown in Table \ref{table3}.  Most results proved Hypothesis 1. \emph{I.e.} the student network tended to encode more visual concepts on the foreground and less concepts on the background, thereby exhibiting a larger ratio $\lambda$ than the baseline network. Figure~\ref{fig4} showed visual concepts encoded in the $ \textrm{FC}_{1}$ layer of VGG-11, which also proved Hypothesis 1.
Note that very few student networks encoded more background visual concepts $N_{\textrm{concept}}^{\textrm{bg}}$. It was because that DNNs used as the teacher network were pre-trained on the ImageNet dataset in Sections \ref{Verification of Hypothesis 1},~\ref{Verification of Hypothesis 2},~\ref{Verification of Hypothesis 3} to verify Hypotheses 1-3. Pre-trained teacher networks encoded visual concepts of $1000$ categories, which were much more than necessary. This would make the student network exhibited a larger $N_{\textrm{concept}}^{\textrm{bg}}$ value than the baseline network.

\subsection{Verification of Hypothesis 2}
\label{Verification of Hypothesis 2}
For Hypothesis 2, we aimed to verify that knowledge distillation enabled the student network to have a higher learning speed, \emph{i.e.} learning different concepts simultaneously. We used $D_{\textrm{mean}}$ and $D_{\textrm{std}}$ to prove this hypothesis.

As shown in Table \ref{table3}, the $D_{\textrm{mean}}$ value and $D_{\textrm{std}}$ value of the student network were both smaller than that of the baseline network, which verified Hypothesis 2. Note that there were still failure cases. For example, the $D_{\textrm{mean}}$ and $D_{\textrm{std}}$ were measured at the  $\textrm{FC}_{1}$ layer of  AlexNet or at the $\textrm{FC}_{2}$ layer of VGG-11. The reason was that AlexNet and VGG-11 both had relatively shallow network architectures. When learning from raw data, DNNs with shallow architectures would learn more concepts and avoid overfitting. Nevertheless, besides very few exceptional cases, knowledge distillation outperformed learning from raw data for most DNNs.

\subsection{Verification of Hypothesis 3}
\label{Verification of Hypothesis 3}
Hypothesis 3 aimed to prove that compared to the baseline network, knowledge distillation made the student network optimized with less detours\textcolor{red}{\footnotemark[1]}. The metric $\rho$ depicted the stability of optimization directions and was used to verify above hypothesis. Results reported in Table \ref{table3} demonstrated that in most cases, the $\rho$ value of the student network was larger than that of the baseline network. When we measured $\rho$ by AlexNet and VGG-11, failure cases emerged due to the shallow architectures of these two networks. Hence, the optimization directions of the student network tended to be unstable and took more detours\textcolor{red}{\footnotemark[1]}.

\begin{table*}[t]
\begin{center}
\resizebox{\linewidth}{!}{\
\settowidth\rotheadsize{\theadfont CUB200-2011 ILSVRC-2013 DET dataset Pascal VOC 2012 dataset}
\begin{tabular}{p{0.05\linewidth}<{\centering} p{0.01\linewidth}<{\centering} p{0.01\linewidth}<{\centering}|p{0.008\linewidth}<{\centering} |p{0.05\linewidth}<{\centering}p{0.05\linewidth}<{\centering} p{0.03\linewidth}<{\centering}| p{0.05\linewidth}<{\centering} p{0.06\linewidth}<{\centering}| p{0.03\linewidth}<{\centering}|p{0.008\linewidth}<{\centering} |p{0.05\linewidth}<{\centering}p{0.05\linewidth}<{\centering} p{0.03\linewidth}<{\centering}| p{0.05\linewidth}<{\centering} p{0.05\linewidth}<{\centering}| p{0.03\linewidth}<{\centering}|p{0.008\linewidth}<{\centering} |p{0.05\linewidth}<{\centering}p{0.05\linewidth}<{\centering} p{0.03\linewidth}<{\centering}| p{0.05\linewidth}<{\centering} p{0.05\linewidth}<{\centering}| p{0.03\linewidth}<{\centering}}
\toprule
\scriptsize{Network} &\scriptsize{Layer}& & &\scriptsize{$N_{\textrm{concept}}^{\textrm{fg}}\uparrow$}& \scriptsize{$N_{\textrm{concept}}^{\textrm{bg}}\downarrow$}&\scriptsize{$\lambda\uparrow$} &\scriptsize{$D_{\textrm{mean}}\downarrow$}& \scriptsize{$D_{\textrm{std}}\downarrow$}& \scriptsize{$\rho\uparrow$}& &\scriptsize{$N_{\textrm{concept}}^{\textrm{fg}}\uparrow$}& \scriptsize{$N_{\textrm{concept}}^{\textrm{bg}}\downarrow$}&\scriptsize{$\lambda\uparrow$} &\scriptsize{$D_{\textrm{mean}}\downarrow$}& \scriptsize{$D_{\textrm{std}}\downarrow$}& \scriptsize{$\rho\uparrow$}& &\scriptsize{$N_{\textrm{concept}}^{\textrm{fg}}\uparrow$}& \scriptsize{$N_{\textrm{concept}}^{\textrm{bg}}\downarrow$}&\scriptsize{$\lambda\uparrow$} &\scriptsize{$D_{\textrm{mean}}\downarrow$}& \scriptsize{$D_{\textrm{std}}\downarrow$}& \scriptsize{$\rho\uparrow$}\\
\cline{1-3} \cline{5-10} \cline{12-17} \cline{19-24}
 \multirow{6}*{\footnotesize{AlexNet}}&\multirow{2}*{\scriptsize{$\textrm{FC}_{1}$}}&\footnotesize{S} &\multirow{28}*{\rothead{CUB200-2011 dataset}} & \textbf{36.60}&  \textbf{4.00}&\textbf{0.90} & 8.35& 25.09& \textbf{0.57}&\multirow{28}*{\rothead{ILSVRC-2013 DET dataset}}&\textbf{49.46} &\textbf{0.66} & \textbf{0.99} & \textbf{0.48}& \textbf{0.10}&\textbf{0.62} &\multirow{28}*{\rothead{Pascal VOC 2012 dataset}}&\textbf{25.84} & \textbf{5.86} & \textbf{0.79} & \textbf{1.14}& \textbf{0.56}& {0.43}\\
& &\footnotesize{B} & &24.13 &5.65 &0.81&\textbf{4.81}& \textbf{14.54}&0.52& &41.00 & 0.92&0.98 & 1.32& 0.31& 0.61& & 20.30 & 6.08&0.77 & 2.00& 2.21& \textbf{0.44}\\
\cline{5-10} \cline{12-17} \cline{19-24}
&\multirow{2}*{\scriptsize{$\textrm{FC}_{2}$}}&\footnotesize{S} & &\textbf{38.13}&  \textbf{3.50}&\textbf{0.92} &\textbf{3.77}& \textbf{3.97}& {0.49} & & \textbf{57.86}& {1.70}& {0.98}& \textbf{0.28}& \textbf{0.01}& {0.60}& & \textbf{31.81}&  {7.29}& \textbf{0.81}& \textbf{0.62}& \textbf{0.07}& \textbf{0.47}\\
& &\footnotesize{B}& &23.33 &5.48 &0.80&5.36& 20.79&\textbf{0.49} & & 42.24& \textbf{0.96} & \textbf{0.98} & 1.15& 0.15& \textbf{0.60}& & 21.85& \textbf{6.56} & 0.77 & 2.04& 1.46& 0.44\\
\cline{5-10} \cline{12-17} \cline{19-24}
&\multirow{2}*{\scriptsize{$\textrm{FC}_{3}$}}&\footnotesize{S} & & \textbf{33.20} &  \textbf{4.31}& \textbf{0.89} & \textbf{8.13} & \textbf{39.79} & \textbf{0.51}&&$-$&$-$ &$-$&$-$&$-$& $-$& &$-$  &$-$& $-$&$-$&$-$ &$-$\\
& &\footnotesize{B}& &22.73 & 4.94 & 0.83& 13.57 & 137.74 & 0.42& &$-$&$-$ &$-$&$-$&$-$&$-$ & & $-$ &$-$&$-$ &$-$&$-$&$-$\\
\cline{1-3} \cline{5-10} \cline{12-17} \cline{19-24}

 \multirow{6}*{\footnotesize{VGG-11}}&\multirow{2}*{\scriptsize{$\textrm{FC}_{1}$}}&\footnotesize{S} & & \textbf{30.69}&  \textbf{10.65}&\textbf{0.75} &\textbf{1.21}& \textbf{0.61}& \textbf{0.56} & & \textbf{44.48}&  \textbf{4.68}&\textbf{0.91} &\textbf{0.26}& \textbf{0.06}& {0.50}& & \textbf{30.56}&  {8.36}&\textbf{0.78} &\textbf{1.09}& \textbf{0.30}& {0.38}\\
& &\footnotesize{B} & &24.26 & 10.77 &0.70 &2.01 & 3.18 &0.55& & 28.27& 7.80&0.80 & 0.93 & 0.08 & \textbf{0.53}& &20.31 & \textbf{7.28}& 0.73&  1.41&  0.54& \textbf{0.44}\\
\cline{5-10} \cline{12-17} \cline{19-24}
&\multirow{2}*{\scriptsize{$\textrm{FC}_{2}$}}&\footnotesize{S} & &\textbf{36.51}& \textbf{10.66} &\textbf{0.78} &\textbf{5.22} & {19.32}& {0.49} & &\textbf{54.20}& \textbf{6.98}&\textbf{0.89} &\textbf{0.18} & \textbf{0.02}& \textbf{0.48}& & \textbf{38.08}& {10.34}& \textbf{0.79}& \textbf{0.70}& \textbf{0.29} & \textbf{0.45}\\
& &\footnotesize{B} & &26.86 &10.71 &0.72 &6.62&\textbf{16.21} &\textbf{0.54}& &29.68 & 8.64& 0.79 & 1.19& 0.52 &0.47& &20.03 & \textbf{7.42}& 0.72& 1.65 & 1.80& 0.36\\
 \cline{5-10} \cline{12-17} \cline{19-24}
&\multirow{2}*{\scriptsize{$\textrm{FC}_{3}$}}&\footnotesize{S} & & \textbf{34.53}&{14.21}& \textbf{0.72} & \textbf{4.15} & \textbf{4.55} & \textbf{0.50} & &$-$&$-$ &$-$&$-$&$-$&$-$ & & $-$ &$-$&$-$ &$-$&$-$&$-$\\
& &\footnotesize{B} & &24.53 & \textbf{10.95} & 0.69 & 20.66& 95.29 & 0.49& &$-$&$-$ &$-$&$-$&$-$&$-$ & & $-$ &$-$&$-$ &$-$&$-$&$-$\\
\cline{1-3} \cline{5-10} \cline{12-17} \cline{19-24}

 \multirow{6}*{\footnotesize{VGG-16}}&\multirow{2}*{\scriptsize{$\textrm{FC}_{1}$}}&\footnotesize{S}  & & \textbf{43.77} & \textbf{8.73} & \textbf{0.84} & \textbf{0.64} & \textbf{0.06} & \textbf{0.66}& & \textbf{56.29}& \textbf{3.13}  &\textbf{0.95} & \textbf{0.02}& \textbf{0.0001}& \textbf{0.47}& & \textbf{42.26} & {11.54} & \textbf{0.80} & \textbf{0.33} & \textbf{0.09} & \textbf{0.52}\\
& &\footnotesize{B} & &22.50 & 11.27 & 0.68 &2.38 & 4.98 & 0.50& &36.06 & 7.71& 0.83& 0.40& 0.13 & 0.44& & 26.87 & \textbf{8.26} & 0.76 & 1.65 & 0.61 & 0.48\\
\cline{5-10} \cline{12-17} \cline{19-24}
&\multirow{2}*{\scriptsize{$\textrm{FC}_{2}$}}&\footnotesize{S}  & &\textbf{36.83} &  \textbf{11.03} & \textbf{0.77} &\textbf{0.80} & \textbf{0.37} & \textbf{0.54}& &{37.79} &  \textbf{4.31}& \textbf{0.90}& \textbf{0.17}& \textbf{0.02}& {0.32}& & \textbf{31.19} & {8.70} & {0.78} & \textbf{0.83} & \textbf{0.45} & {0.35}\\
& &\footnotesize{B} & &23.31 &11.56 &0.67 &5.43 & 22.96 &0.50& &\textbf{38.41} & 9.66 &0.80& 0.79& 0.52&\textbf{0.43}& & 29.37 & \textbf{8.04} & \textbf{0.78} & 2.65 & 1.90 & \textbf{0.46}\\
\cline{5-10} \cline{12-17} \cline{19-24}
&\multirow{2}*{\scriptsize{$\textrm{FC}_{3}$}}&\footnotesize{S}  & & \textbf{32.32} & {10.21} & \textbf{0.77} & \textbf{6.17} & \textbf{32.63} & \textbf{0.47}& &$-$&$-$ &$-$&$-$&$-$&$-$ & & $-$ &$-$&$-$ &$-$&$-$&$-$\\
& &\footnotesize{B} & &23.26 & \textbf{9.97} & 0.71 &17.53 & 216.05 & 0.46& &$-$&$-$ &$-$&$-$&$-$&$-$ & & $-$ &$-$&$-$ &$-$&$-$&$-$\\
\cline{1-3} \cline{5-10} \cline{12-17} \cline{19-24}

\multirow{6}*{\footnotesize{VGG-19}}&\multirow{2}*{\scriptsize{$\textrm{FC}_{1}$}}&\footnotesize{S}  & & \textbf{40.74}&  \textbf{10.42}& \textbf{0.80} &\textbf{0.66} & \textbf{0.15} & \textbf{0.60}& & \textbf{46.50}&  \textbf{2.52}& \textbf{0.95}&\textbf{0.16} & \textbf{0.0002}& \textbf{0.39}& &\textbf{46.38}& {14.05}& {0.77} & \textbf{0.25}& \textbf{0.07}& \textbf{0.45} \\
& &\footnotesize{B} & &22.42 &11.19 & 0.67 & 2.33 &3.67 &0.47& &29.71 &5.83& 0.84& 0.33& 0.12 &0.39& & 28.65&  \textbf{7.93}& \textbf{0.78}& 1.10& 0.80& 0.41\\
\cline{5-10} \cline{12-17} \cline{19-24}
&\multirow{2}*{\scriptsize{$\textrm{FC}_{2}$}}&\footnotesize{S}  & & \textbf{40.20} & \textbf{9.03} & \textbf{0.82} & \textbf{1.16} & \textbf{0.63} & \textbf{0.56}& & \textbf{50.90}& \textbf{5.96} & \textbf{0.91} & \textbf{0.06}& \textbf{0.0006}& \textbf{0.37}& & \textbf{47.03}& {13.66}  & \textbf{0.78}& \textbf{0.10} & \textbf{0.03} & \textbf{0.45}\\
& &\footnotesize{B} & &24.00 &10.40 & 0.70 & 4.64 & 19.07 & 0.47& & 30.31 & 6.15 & 0.84 & 0.45 & 0.18 & 0.37& & 28.46& \textbf{8.20} & 0.78 & 2.14 & 1.92 & 0.41\\
\cline{5-10} \cline{12-17} \cline{19-24}
&\multirow{2}*{\scriptsize{$\textrm{FC}_{3}$}}&\footnotesize{S}  & &\textbf{28.60} &  \textbf{6.37}& \textbf{0.82}& \textbf{4.89}& \textbf{11.57}& \textbf{0.48} & &$-$&$-$ &$-$&$-$&$-$&$-$ & & $-$ &$-$&$-$ &$-$&$-$&$-$\\
& &\footnotesize{B} & &21.29 & 7.77 & 0.74 & 20.61 &143.61 & 0.46& &$-$&$-$ &$-$&$-$&$-$&$-$ & & $-$ &$-$&$-$ &$-$&$-$&$-$\\
\cline{1-3} \cline{5-10} \cline{12-17} \cline{19-24}

\multirow{6}*{\scriptsize{ResNet-50}}&\multirow{2}*{\scriptsize{$\textrm{FC}_{1}$}}&\footnotesize{S} & & \textbf{43.02}&  \textbf{10.15}&\textbf{0.81} & 24.43& 166.76& 0.48& & \textbf{56.00}&  {6.50}&\textbf{0.90} & 3.45& \textbf{4.74}& \textbf{0.45}& & \textbf{42.54}& {10.76}&0.80 & 3.43& 19.60& \textbf{0.40}\\
& &\footnotesize{B} & &42.15 &11.83 &0.79&\textbf{20.78}& \textbf{122.79}&\textbf{0.53}& &43.80 &\textbf{5.75} &0.89&\textbf{2.73}& 6.82&0.36& &39.65 &\textbf{9.81} &\textbf{0.81}&\textbf{1.64}& \textbf{15.20}&0.39\\
\cline{5-10} \cline{12-17} \cline{19-24}
&\multirow{2}*{\scriptsize{$\textrm{FC}_{2}$}}&\footnotesize{S} & &\textbf{48.58}&  \textbf{9.75}&\textbf{0.83} &37.62& \textbf{206.22}& \textbf{0.55}& &\textbf{52.57}&  \textbf{6.54}&\textbf{0.90} &0.25& {1.45}& \textbf{0.40}& &\textbf{41.03}& {12.37}&{0.77} &\textbf{1.85}& \textbf{13.03}& \textbf{0.41}\\
& &\footnotesize{B} & &42.06 &11.88 &0.79&\textbf{29.28}& 248.03&0.52& &43.63 &6.93 &0.87&\textbf{0.02}& \textbf{0.02}&0.35& &38.00 &\textbf{10.00} &\textbf{0.80}&2.68& 30.91&0.38\\
\cline{5-10} \cline{12-17} \cline{19-24}
&\multirow{2}*{\scriptsize{$\textrm{FC}_{3}$}}&\footnotesize{S}  & & {41.38}  &{11.73}  &0.77  &926.61  &\scriptsize{142807.00}  &0.43 & &$-$&$-$ &$-$&$-$&$-$&$-$ & & $-$ &$-$&$-$ &$-$&$-$&$-$\\
& &\footnotesize{B} & &\textbf{42.03}  &\textbf{11.48}  &\textbf{0.79} &\textbf{111.18}  &\textbf{3299.20}  &\textbf{0.53}& &$-$&$-$ &$-$&$-$&$-$&$-$ & & $-$ &$-$&$-$ &$-$&$-$&$-$\\
\cline{1-3} \cline{5-10} \cline{12-17} \cline{19-24}

\multirow{6}*{\scriptsize{ResNet-101}}&\multirow{2}*{\scriptsize{$\textrm{FC}_{1}$}}&\footnotesize{S} & & \textbf{45.93}&  \textbf{11.14}&\textbf{0.81} & \textbf{23.32}& \textbf{236.76}& 0.51& & \textbf{48.59}&  \textbf{5.06}&\textbf{0.91} & \textbf{1.99}& \textbf{2.20}& \textbf{0.39}& & {42.54}&  {9.37}&{0.82} & \textbf{1.39}& \textbf{32.87}& {0.35}\\
& &\footnotesize{B} & &44.18 &12.55 &0.78&40.41& 828.72&\textbf{0.52}& &42.94 &8.16 &0.84&{5.41}& 10.39&0.35& &\textbf{43.33} &\textbf{9.30} &\textbf{0.83}&{15.28}& {48.71}&\textbf{0.39}\\
\cline{5-10} \cline{12-17} \cline{19-24}
&\multirow{2}*{\scriptsize{$\textrm{FC}_{2}$}}&\footnotesize{S} & &\textbf{51.59}&  \textbf{9.02}&\textbf{0.85} &67.60& \textbf{947.85}& \textbf{0.54}& &\textbf{49.27}&  \textbf{6.39}&\textbf{0.89} &\textbf{0.98}& \textbf{0.65}& \textbf{0.37}& &\textbf{41.71}&  {9.16}&{0.82} &{3.30}& {100.97}& {0.38}\\
&&\footnotesize{B} & &43.22 &12.32 &0.78&\textbf{43.40}& \small{1155.22}&0.50& &41.79 &7.30 &0.85&{6.58}& {17.16}&0.34& &41.35 &\textbf{8.32} &\textbf{0.84}&\textbf{2.26}& \textbf{48.61}&\textbf{0.39}\\
 \cline{5-10} \cline{12-17} \cline{19-24}
&\multirow{2}*{\scriptsize{$\textrm{FC}_{3}$}}&\footnotesize{S}  & &\textbf{47.71}  &\textbf{10.24}  &\textbf{0.82}  &\textbf{73.33}  &\textbf{2797.15}  &\textbf{0.53}  & &$-$&$-$ &$-$&$-$&$-$&$-$ & & $-$ &$-$&$-$ &$-$&$-$&$-$\\
& &\footnotesize{B} & &42.40  &10.53  &{0.80} &{162.68}  &\footnotesize{16481.93}  &0.49& &$-$&$-$ &$-$&$-$&$-$&$-$ & & $-$ &$-$&$-$ &$-$&$-$&$-$\\
\cline{1-3} \cline{5-10} \cline{12-17} \cline{19-24}

\multirow{6}*{\scriptsize{ResNet-152}}&\multirow{2}*{\scriptsize{$\textrm{FC}_{1}$}}&\footnotesize{S} & & {44.81}& {12.09}&{0.79} & \textbf{26.35}& \textbf{289.59}& 0.48& & \textbf{44.90}&  {5.63}&\textbf{0.89} & {6.25}& \textbf{5.86}& \textbf{0.36}& & \textbf{41.09}&  \textbf{10.09}&\textbf{0.81} & \textbf{0.33}& \textbf{3.59}& \textbf{0.39}\\
& &\footnotesize{B} & &\textbf{45.62} &\textbf{10.68} &\textbf{0.81}&36.92& 767.58&\textbf{0.54}& &39.93 &\textbf{5.40} &0.89&\textbf{6.08}& 6.74&0.33& &40.15 &10.82 &{0.79}&{0.59}& {11.39}&{0.37}\\
\cline{5-10} \cline{12-17} \cline{19-24}
&\multirow{2}*{\scriptsize{$\textrm{FC}_{2}$}}&\footnotesize{S} & &{43.79}&  \textbf{10.04}&\textbf{0.81} &\textbf{7.13}& \textbf{42.77}& {0.52}& &\textbf{40.98}&  {6.90}&{0.86} &\textbf{4.64}& \textbf{5.71}& {0.32}& &\textbf{41.36}& \textbf{12.04}&\textbf{0.78} &\textbf{14.29}& \textbf{17.33}& \textbf{0.38}\\
&&\footnotesize{B} & &\textbf{45.08} &10.85 &0.81&{44.59}& \small{1200.97}&\textbf{0.52}& &{40.29} &\textbf{5.56} &\textbf{0.89}&{7.86}& {12.24}&\textbf{0.33}& &38.57 &12.07 &{0.77}&{18.03}& {67.52}&{0.36}\\
\cline{5-10} \cline{12-17} \cline{19-24}
&\multirow{2}*{\scriptsize{$\textrm{FC}_{3}$}}&\footnotesize{S}  & &{44.21}  &{11.89}  &{0.79}  &\textbf{47.28}  &\textbf{1463.55}  &{0.50}& &$-$&$-$ &$-$&$-$&$-$&$-$ & & $-$ &$-$&$-$ &$-$&$-$&$-$\\
& &\footnotesize{B} & &\textbf{44.89}  &\textbf{10.77}  &\textbf{0.81} &{167.41}  &\footnotesize{16331.28}  &\textbf{0.52}& &$-$&$-$ &$-$&$-$&$-$&$-$ & & $-$ &$-$&$-$ &$-$&$-$&$-$\\

\bottomrule
\end{tabular}}
\end{center}\vspace{-5pt}
\caption{Comparisons between the student network (S) and the baseline network (B). $\uparrow$/$\downarrow$ indicates that larger/smaller values were better. In general, the student network had larger values of $N_{\textrm{concept}}^{\textrm{fg}}$, $\lambda$, $\rho$, and smaller values of $N_{\textrm{concept}}^{\textrm{bg}}$, $D_{\textrm{mean}}$, $D_{\textrm{std}}$ than the baseline network, which proved Hypotheses 1-3.}
\label{table3}
\vspace{-15pt}
\end{table*}

\section{Conclusion and Discussions}
In this paper, we interpret the success of knowledge distillation from the perspective of quantifying the knowledge encoded in the intermediate layer of a DNN. Three types of metrics were proposed to verify three hypotheses in the scenario of classification. \emph{I.e.} knowledge distillation ensures DNNs learn more task-relevant concepts and less task-irrelevant concepts, have a higher learning speed, and optimize with less detours\textcolor{red}{\footnotemark[1]} than learning from raw data.

There are several limitations of our work. We only focus on the classification task in this paper. However, applying our methods to other tasks (\emph{e.g.} object segmentation), or other types of data (\emph{e.g.} the video) is theoretically feasible. Meanwhile, for these tasks, side information may be required. In this paper, our proposed metrics are implemented by using the entropy-based analysis, which has strong connections to the information-bottleneck theory. Unlike the information-bottleneck theory, the proposed metrics can measure the pixel-wise discarding. However, the learning procedure of DNNs cannot be precisely divided into the learning phase and the discarding phase. In each epoch, the DNN may simultaneously learn new visual concepts and discard old task-irrelevant concepts. Thus, the target epoch $\hat{m}$ in Figure~\ref{fig2} is just a rough estimation of the division of two learning phases.

\section*{Acknowledgements}
The corresponding author Quanshi Zhang is with the John Hopcroft Center and MoE Key Lab of Artificial Intelligence AI Institute, Shanghai Jiao Tong University. He thanks the support of National Natural Science Foundation of China (U19B2043 and 61906120) and Huawei Technologies. Zhefan Rao and Yilan Chen made equal contribution to this work as interns at Shanghai Jiao Tong University.

\bibliographystyle{ieee_fullname}
\bibliography{v16}

\end{document}